\documentclass[letterpaper]{article} 
\usepackage{aaai23}  
\usepackage{times}  
\usepackage{helvet}  
\usepackage{courier}  
\usepackage[hyphens]{url}  
\usepackage{graphicx} 
\usepackage{natbib}  
\usepackage{caption} 
\usepackage{algorithm}
\usepackage{algorithmic}
\graphicspath{{images/}}
\usepackage{newfloat}
\usepackage{listings}
\DeclareCaptionStyle{ruled}{labelfont=normalfont,labelsep=colon,strut=off} 
\floatstyle{ruled}
\newfloat{listing}{tb}{lst}{}
\floatname{listing}{Listing}
\usepackage{amsmath}
\usepackage{amsthm}
\usepackage{amssymb}
\usepackage{bm}
\usepackage{multirow}

\usepackage{color}
\usepackage{booktabs}
\usepackage{rotating}
\usepackage{subfig}

\title{Incomplete Multi-View Multi-Label Learning via Label-Guided Masked View- and Category-Aware Transformers}
\author {
        Chengliang Liu\textsuperscript{\rm 1},
        Jie Wen\textsuperscript{\rm 1$\ast$},
        Xiaoling Luo\textsuperscript{\rm 1},
        Yong Xu\textsuperscript{\rm 1,\rm 2}\thanks{Corresponding authors.}
}
\affiliations {
    \textsuperscript{\rm 1} Shenzhen Key Laboratory of Visual Object Detection and Recognition, Harbin Institute of Technology, Shenzhen, China\\
    \textsuperscript{\rm 2} Pengcheng Laboratory, Shenzhen, China.\\
    liucl1996@163.com, jiewen\_pr@126.com, xiaolingluoo@outlook.com, yongxu@ymail.com  
}

\usepackage{bibentry}

\begin{document}

\maketitle

\begin{abstract}
As we all know, multi-view data is more expressive than single-view data and multi-label annotation enjoys richer supervision information than single-label, which makes multi-view multi-label learning widely applicable for various pattern recognition tasks. In this complex representation learning problem, three main challenges can be characterized as follows: i) How to learn consistent representations of samples across all views? ii) How to exploit and utilize category correlations of multi-label to guide inference? iii) How to avoid the negative impact resulting from the incompleteness of views or labels? To cope with these problems, we propose a general multi-view multi-label learning framework named label-guided masked view- and category-aware transformers in this paper. First, we design two transformer-style based modules for cross-view features aggregation and multi-label classification, respectively. The former aggregates information from different views in the process of extracting view-specific features, and the latter learns subcategory embedding to improve classification performance. Second, considering the imbalance of expressive power among views, an adaptively weighted view fusion module is proposed to obtain view-consistent embedding features. Third, we impose a label manifold constraint in sample-level representation learning to maximize the utilization of supervised information. Last but not least, all the modules are designed under the premise of incomplete views and labels, which makes our method adaptable to arbitrary multi-view and multi-label data. Extensive experiments on five datasets confirm that our method has clear advantages over other state-of-the-art methods.
\end{abstract}

\section{Introduction}

Since the development of representation learning methods, data analysis technology based on simple single-view data has become more and more difficult to meet diverse application requirements. In the past few years, data acquisition technology has flourished, and multi-view data from various media or with different styles are ubiquitous, providing more possibilities for comprehensively and accurately describing observation targets. Simply put, multiple observations,  obtained from different observation angles for the same thing, could be regarded as multi-view data \cite{li2020bipartite,li2021incomplete,wang2022mvsnet}. For example, retinal images captured at four different positions constitute a four-view retinal dataset \cite{luo2021mvdrnet,wen2022survey,luo2023deep}; the features extracted from the target image with different feature extraction operators can also form a multi-view dataset \cite{liu2022localized,hu2021view}. More typically, multi-view or multi-modal datasets composed of multimedia data such as text, pictures, and videos have been widely used in many applications such as web page classification \cite{hu2020dmib,lu2019online,liu2023dicnet}. As a result, multi-view learning emerges as the times require, and a large number of methods based on subspace learning, matrix factorization, and graph learning have been proposed \cite{zhan2017graph,liu2017optimal,wen2019unified}. Most of these methods seek to obtain a consistent representation of multiple views to characterize the essential attributes of objects. On the other hand, since the core of multi-view learning is representation learning, researchers usually combine it with downstream tasks to improve the application value and evaluate the learning effect \cite{chen2022low}. That is, multi-view learning can be divided into clustering and classification, according to whether supervised information is available. Further, in the single-label classification task, a sample is only labeled with one category, which is obviously against the distribution of information in nature \cite{zhao2022shared}. For example, a bird picture is likely to contain the category of `sky' or `tree'. Therefore, although multi-label classification still faces more challenges than single-label classification, its broad application prospects are attracting a lot of research enthusiasm. Based on this, a complex fusion task, \textit{i.e.}, multi-view multi-label classification (\textit{MvMlC}) is put on the agenda. \citeauthor{zhang2018latent} proposed a matrix factorization-based \textit{MvMlC} model, which attempts to enforce alignment of different views in the kernel space to exploit multi-view complementarity and explore more latent information \cite{zhang2018latent}. A Bernoulli mixture conditional model is proposed to model label dependencies and employ a variational inference framework for approximate posterior estimation \cite{sun2020lcbm}. 

Researchers have conducted extensive researches in the field of \textit{MvMlC}, however these works assume that all views and labels are complete, which is often violated in practice. For example, if a web page contains only text and images, then the video view of the page is not available. To avoid the negative effects of missing views as much as possible, some multi-view learning works try to mask unavailable views or restore missing views \cite{wen2022survey,xu2018partial}. Similarly, manual annotation is likely to miss some tags due to mistakes or cost, which inevitably weakens the supervision information of multi-label. To solve the issue, some works focusing on incomplete multi-label classification have been developed in recent years \cite{zhu2017multi,huang2019improving}. Although these methods designed for incomplete multi-view or incomplete multi-label learning have achieved surprising fruits, most of them are not able to cope with the both incomplete cases simultaneously. 

To this end, we propose a general \textit{MvMlC} framework, termed \textbf{L}abel-guided \textbf{M}asked \textbf{V}iew- and \textbf{C}ategory-\textbf{A}ware \textbf{T}ransformers (\textbf{LMVCAT}), which can handle the multi-view data with arbitrary missing-view and missing-label. In addition, unlike traditional methods to learn low-level representations of samples, deep neural networks based LMVCAT is capable to extract high-level features of samples, which is of great benefit for learning complex multi-category semantic labels. In recent years, transformer, designed for natural language processing, has shown its dominance in the image and other fields, which clearly demonstrates the effectiveness of the self-attention mechanism \cite{vaswani2017attention,huang2022pixel,huang2022weakly}. Inspired by this, our LMVCAT is also designed based on the transformer with self-attention mechanism which can provide a global receptive field for each view \cite{lanchantin2021general}. Overall, our model is composed of four parts: a masked view-aware encoder, an adaptively weighted multi-view fusion module, a label-guided sample-level graph constraint, and a subcategory-aware multi-label classification module. Specifically, the four parts of our LMVCAT are motivated by the following points: 1) breaking the inter-view barriers to aggregate multi-view features can fully exploit the complementary information of multiple views, so we design a view-aware module to integrate all views while extracting sample-specific high-level representations; 2) different discrimination ability of views means the different contributions to the final classification, so an adaptively weighted strategy that automatically learns the weight factor of each view is needed \cite{chen2022adaptively}; 3) compared with single-label data, multi-label data naturally enjoys richer label similarity information, so it is of great significance to utilize label smoothness (manifold) to guide sample encoding; 4) it is well known that multi-label data is with non-negligible inter-class correlations, thus we adopt a category-aware module that learns correlations in the subcategory embedding space to collaboratively predict labels \cite{lanchantin2021general}. Apart from those, it needs to be emphasized that we consider the possibility of missing labels and missing views in all components of our model. So it is no doubt that our LMVCAT is a general \textit{MvMlC} framework that is comfortable with all kinds of multi-view multi-label data.
Our contributions are summarized as follows:
\begin{itemize}
    \item To the best of our knowledge, this is the first Transformer-based incomplete multi-view multi-label learning framework capable of handling both incomplete views and labels. The proposed masked view-aware self-attention module could avoid the missing views' negative effects to information interaction across views. Similarly, our category-aware module is designed to mine latent inter-class correlations in the subcategory embedding space to improve the expressiveness.
    \item Label manifold is fully exploited. Although the multi-label information is fragmentary, we still try our best to construct an approximate similarity graph based on incomplete labels to guide the encoding process of samples, which further strengthens the discrimination power of high-level semantic features.
    \item Different views contribute different importance to the prediction, so we introduce an adaptively weighted fusion method to balance this importance instead of simply adding multiple views. Sufficient experimental results confirm the effectiveness of our LMVCAT.
\end{itemize}

\section{Preliminaries}
\subsection{Formulation}
In this section, we define our main problem as follows: Given input multi-view dataset $\bm{\mathsf{D}}= \{\bm{\mathsf{X}},\bm{\mathsf{Y}}\}$, which contains $n$ samples. For $i$-th sample, $\bm{\mathsf{X}}_{i}$ is composed of $m$ views with dimension $d_v$, \textit{i.e.}, $\bm{\mathsf{X}}_{i}=\bigl\{x_{i}^{(v)}\in \mathbb{R}^{d_{v}}\bigr\}_{v=1}^{m}$, or for $v$-th view, $\bm{\mathsf{X}}^{(v)}=\bigl\{x^{(v)}_{i}\in \mathbb{R}^{d_{v}}\bigr\}_{i=1}^{n}$. $\bm{\mathsf{Y}}_{i}\in \{0,1\}^{c}$ is a row vector that denotes the label of $i$-th sample and $c$ is the number of categories. To describe the missing cases, we let $\bm{\mathsf{W}}\in \{0,1\}^{n\times m}$ be the missing-view indicator matrix, where $\bm{\mathsf{W}}_{ij}=1$ represent $j$-th view of sample $i$ is available, otherwise $\bm{\mathsf{W}}_{ij}=0$. Similarly, we define $\bm{\mathsf{G}}\in \{0,1\}^{n\times c}$ as the missing-label indicator matrix, where $\bm{\mathsf{G}}_{ij}=1$ represents $j$-th category of sample $i$ is known, otherwise $\bm{\mathsf{G}}_{ij}=0$. All missing information of feature $\bm{\mathsf{X}}$ and label $\bm{\mathsf{Y}}$ will be set as `0' in the data-preparation stage. And our goal of multi-view multi-label learning is to train a model which can correctly predict multiple categories for each input sample. 
\begin{figure*}[h!]
\centering
\includegraphics[width=0.9\textwidth]{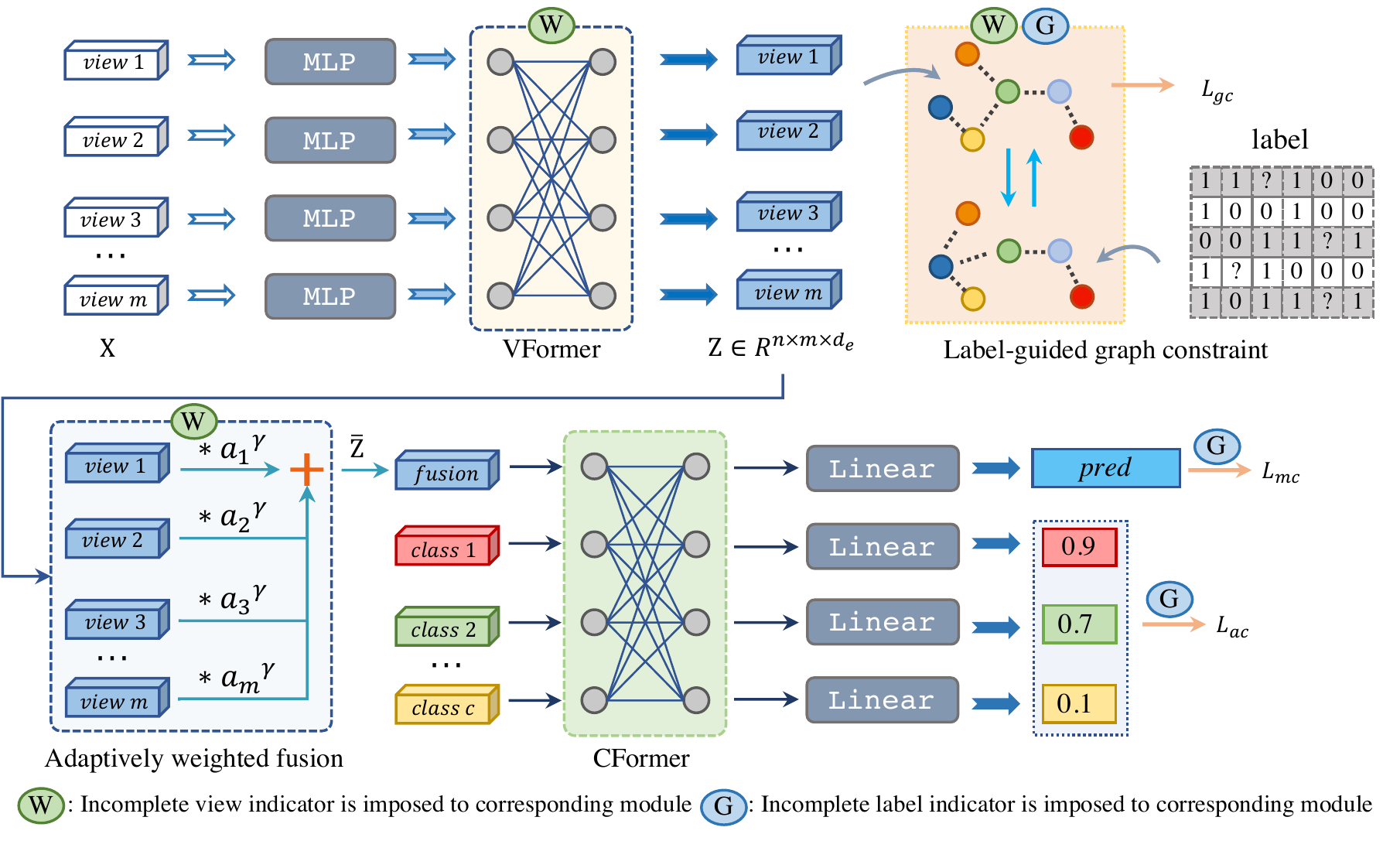} 
\caption{The main structure of our LMVCAT. The MLP layer is composed of two linear layers, GELU activation function, and two dropout layers. And \textit{class} 1 to \textit{class} c denote the $c$ category tokens. Incomplete view indicator and incomplete label indicator are imported to corresponding modules and losses.}
\label{fig:main}
\end{figure*}

\subsection{Related Works}
\label{2.2}
A matrix completion based \textit{MvMlC} method, termed as lrMMC, which forces common subspace to be low rank to satisfy the assumption of matrix completion \cite{liu2015low}. However, lrMMC is incapable of adapting to missing-view or missing-label data. MVL-IV, an incomplete multi-view learning method, attempts to exploit the connections between views \cite{xu2015multi}. As another incomplete multi-view single-label learning model, iMSF cleverly divides the incomplete multi-view classification tasks into multiple complete subtasks \cite{yuan2012multi}. There is a limitation to both methods that MVL-IV and iMSF only consider the incompleteness of views. On the contrary, MvEL, a method aiming to capture the semantic context and the neighborhood consistency, could only handle the incomplete multi-label case \cite{zhang2013multi}. In recent years, there are few approaches to consider double missing cases except iMvWL\cite{tan2018incomplete} and NAIML\cite{li2020concise}. iMvWL simultaneously maps multi-view features and muti-label information to a common subspace. And a correlation matrix is introduced to enhance the projection from label space to embedding subspace. NAIML copes with the double incomplete problem based on the low-rank hypothesis of sub-label matrix, which also implicitly exploits the sub-class correlations. In addition, because the iMvWL and NAIML are well suited to datasets with both view and label incompleteness, we focus on evaluating these two methods in our comparison experiments.

\section{Method}
In this section, we elaborate on each component of our model, including view-aware transformer, label-guided graph constraint, multi-view adaptively weighted fusion, and category-aware transformer.
\subsection{VFormer}
\begin{figure}[t]
\centering
\includegraphics[width=0.9\columnwidth]{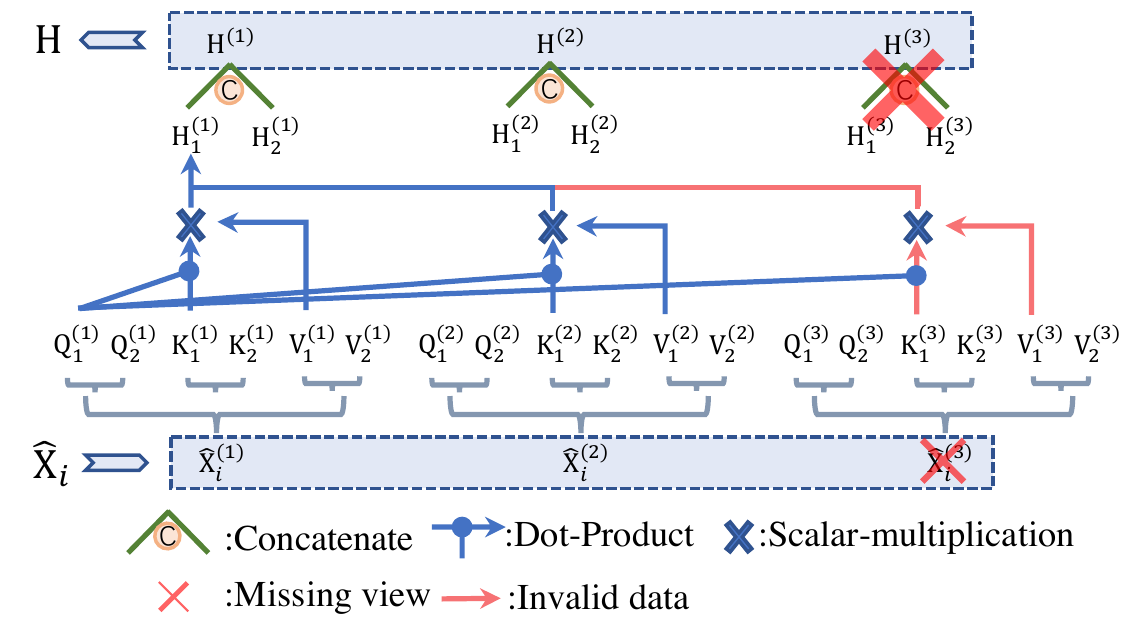} 
\caption{The masked multi-head self-attention mechanism of VFormer taking two heads as an example.}
\label{figure:attention}
\end{figure}

As we all know, the key to success of multi-view learning is the complementarity among views, which are not available in single-view data. To this end, we design a view-aware transformer encoder (\textbf{VFormer} for short) to aggregate complementary information during the cross-view interaction. Before this, it needs to be considered that different views may have different feature dimensions, so for convenience, we map the original multi-view data to the embedding space with the same dimension by a stack of Multilayer Perceptrons (MLP) $\big\{\bm{\varPhi}_\theta^{(v)}\big\}_{v=1}^{m}$, which can also be seen as a preliminary feature extraction operation, \textit{i.e.}, $\bm{\varPhi}_\theta^{(v)}: \bm{\mathsf{X}}^{(v)}\in \mathbb{R}^{n\times d_v} \rightarrow \bm{\mathsf{\widehat{X}}}^{(v)}\in \mathbb{R}^{n\times d_e}$. These embedding features of multiple views are stacked into a feature sequence $\bm{\mathsf{\widehat{X}}}\in \mathbb{R}^{n\times m \times d_e}$, like a sentence embedding tensor composed of some word embedding vectors, as the input tensor of the VFormer. The structure of our VFormer is similar to that of the encoder in the typical transformer \cite{vaswani2017attention}, and the main difference is that we introduce a missing view indicator matrix in the calculation of multi-head self-attention scores to prevent missing views from participating in the calculation of attention scores. Our masked multi-head self-attention encoder is characterized as follows:

For each sample embedding $\bm{\mathsf{\widehat{X}}}_i \in \mathbb{R}^{m\times d_e}$, we project it linearly to get its queries, keys, and values by $h$ groups projective matrices, \textit{i.e.}, $\{\bm{\mathsf{W^q}}_t,\bm{\mathsf{W^k}}_t,\bm{\mathsf{W^v}}_t\}_{t=1}^{h}$ with head number $h$. To mask the embedding features according to missing-view distribution, we define a \textit{fill} function to fill zero value with $-1e^9$ and construct a mask matrix of sample $i$: $\bm{\mathsf{M}}_i =w_i^Tw_i \in \mathbb{R}^{m\times m}$, where $w_i$ is $i$-th row vector of $\bm{\mathsf{W}}$. Then we calculate view correlations $\bm{\mathsf{A}}_t$ and output $\bm{\mathsf{H}}_t$:
\begin{equation}
\small
\begin{aligned}
&\bm{\mathsf{A}}_t=softmax\Bigl(fill\bigl(\bigl(\bm{\mathsf{\widehat{X}}}_i\bm{\mathsf{W^q}}_t\bigr)\bigl(\bm{\mathsf{\widehat{X}}}_i\bm{\mathsf{W^k}}_t\bigr)^{T}\bm{\mathsf{M}}_i\bigr)/\sqrt{d_h}\Bigr)\\
& \bm{\mathsf{H}}_t = \bm{\mathsf{A}}_t\bigl(\bm{\mathsf{\widehat{X}}}_i\bm{\mathsf{W^v}}_t\bigr),
\end{aligned}
\end{equation}	
where $d_h=d_e/h$, $\bm{\mathsf{W^q}}_t, \bm{\mathsf{W^k}}_t, \bm{\mathsf{W^v}}_t \in \mathbb{R}^{d_e \times d_h}$, and $\bm{\mathsf{H}}_t\in \mathbb{R}^{m\times d_h}$. The masked self-attention mechanism is shown in the Fig. \ref{figure:attention}, it is worth noting that we fill the attention values with $-1e^9$ so that the softmax will ignore the corresponding missing views when calculating the attention scores. And then, we concatenate all outputs to produce a new embedding feature \textit{w.r.t} sample $i$:
$\bm{\mathsf{H}} = Concat(\bm{\mathsf{H}}_1,...,\bm{\mathsf{H}}_t) \in \mathbb{R}^{m\times d_e}$. To sum up, all views of the same sample will exchange information during the parallel encoding process in our VFormer. As a result, the private information of each view is shared to some extent by other views. Other operations on VFormer are shown in the Fig. \ref{figure:2fig}(a). Finally, our VFormer can be formulated as: $\bm{\varGamma}:\bm{\mathsf{\widehat{X}}}\rightarrow\bm{\mathsf{Z}}\in \mathbb{R}^{n\times m\times d_e}$.

\subsection{Label-Guided Graph Constraint}
\begin{figure}[t]
\centering
\includegraphics[width=0.9\columnwidth]{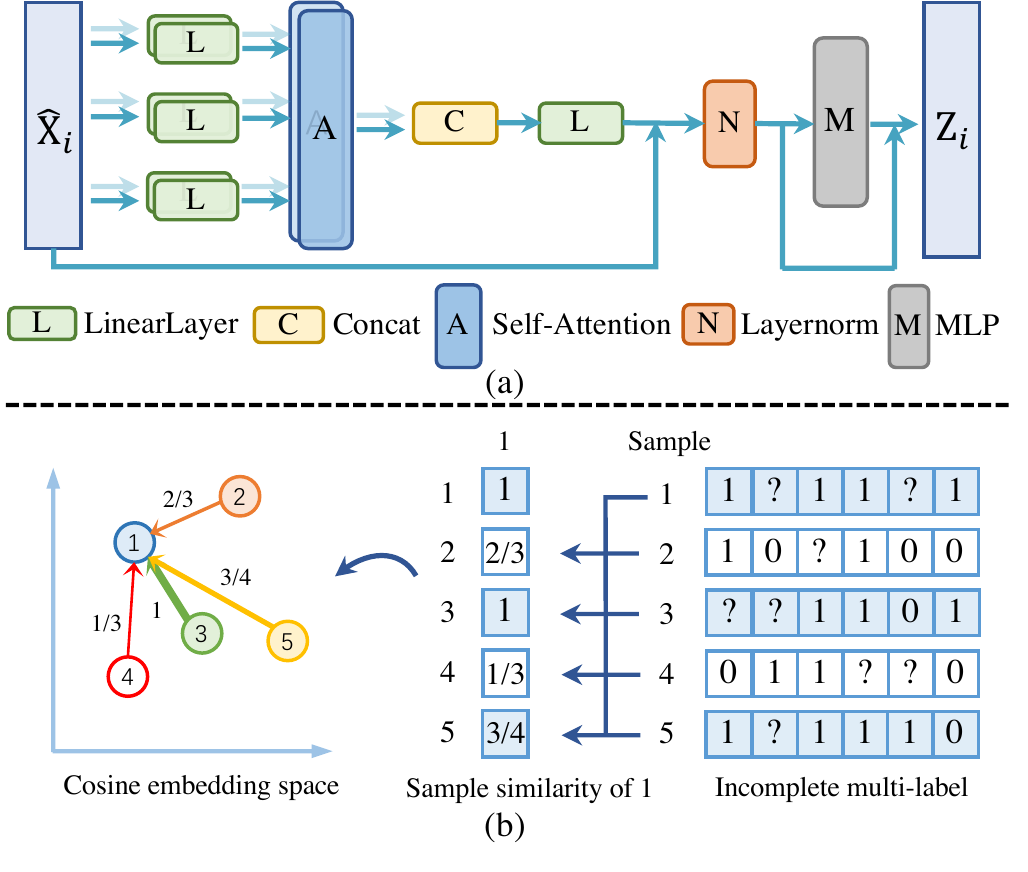} 
\caption{(a) The brief construction of VFormer taking sample $i$ as an example; (b) The simple schematic diagram of label-guided graph constraint.}
\label{figure:2fig}
\end{figure}

Unlike single-label samples maintaining an invariant label distance ($\sqrt{2}$ by Euclidean distance), multi-label samples naturally hold uneven distribution in label space, which offers the possibility to guide the high-level representation learning based on label similarity. Simply put, the label manifold assumption means that if two samples are similar, their labels should also be similar \cite{wu2014multi}. In turn, we utilize label similarity to construct a sample-level graph constraint to guide the representation learning. As shown in Fig. \ref{figure:2fig}(b), the similarity vector from sample 1 to other samples is calculated to guide the embedding encoding of sample 1. Before that, we define the label similarity matrix $\bm{\mathsf{T}}$:
\begin{equation}
\small
\bm{\mathsf{T}}=\bigl(\bm{\mathsf{Y}}\odot\bm{\mathsf{G}}\bigr)\bigl(\bm{\mathsf{Y}}\odot\bm{\mathsf{G}}\bigr)^T./\bigl(\bm{\mathsf{G}}\bm{\mathsf{G}}^T\bigr),
\end{equation}
where ‘$\odot$’ is Hadamard product and ‘$./$’ denotes the division of corresponding elements. Notably, $\bm{\mathsf{T}}\in [0,1]^{n\times n}$ is normalized by $\bm{\mathsf{G}}\bm{\mathsf{G}}^T$, where $(\bm{\mathsf{G}}\bm{\mathsf{G}}^T)_{ij}$ is referring to the number of known categories in both $\bm{\mathsf{Y}}_i$ and $\bm{\mathsf{Y}}_j$. In other words, for two samples, the larger the number of common positive tags, the more similar they are. In addition, the similarity of two embedding features is calculated in cosine space, \textit{i.e.}, for sample $i$ and $j$, their similarity in view $v$ is defined as:
\begin{equation}
\small
\bm{\mathsf{S}}_{ij}^{(v)}=(\frac{\big\langle z_{i}^{(v)} \cdot z_{j}^{(v)}\big\rangle}{\big\|z_{i}^{(v)}\big\| \big\|z_{j}^{(v)}\big\|}+1)/2,
\end{equation}
where $\langle\cdot\rangle$ denotes the vector dot product operation. $z_{i}^{(v)}$ and $z_{j}^{(v)}$ are two embedding features from view $v$ that are output by VFormer. In order to learn sample neighbor relationships from labels, we let label similarity be the target and feature similarity be the learning object. The graph constraint loss $\bm{\mathcal{L}}_{gc}$ is formulated as:
\begin{equation}
\small
\begin{aligned}
\bm{\mathcal{L}}_{gc} &= -\frac{1}{2mN}\sum_{v=1}^{m}\sum_{i=1}^{n}\sum_{j\ne i}^{n}\bigl(\bm{\mathsf{T}}_{ij}\log{\bm{\mathsf{S}}^{(v)}_{ij}}\\
&+\bigl(1-\bm{\mathsf{T}}_{ij}\bigr)\log{\bigl(1-\bm{\mathsf{S}}^{(v)}_{ij}\bigr)}\bigr)(\bm{\mathsf{W}}_{iv}\bm{\mathsf{W}}_{jv}),
\end{aligned}
\end{equation}
where $N=\sum_{i,j}{\bm{\mathsf{W}}_{iv}\bm{\mathsf{W}}_{jv}}$ denotes the number of available sample pairs in view $v$. We introduce $\bm{\mathsf{W}}_{iv}\bm{\mathsf{W}}_{jv}$ to mask the calculation of loss \textit{w.r.t} missing views.

\subsection{Adaptively Weighted Fusion}
As mentioned above, VFormer encodes an embedding feature for each view of each sample (\textit{i.e.}, $\bm{\mathsf{Z}}\in \mathbb{R}^{n\times m\times d_e}$). In order to obtain a consistent common representation to uniquely describe the corresponding sample, we propose an adaptively weighted fusion strategy to fuse multi-view embedding features before final classification. The fusion feature $\bar{z_i}$ of sample $i$ is defined as follows:

\begin{equation}
\small
\bar{z_i}=\sum_{v=1}^{m}\frac{e^{a^{\gamma}_{v}}z_{i}^{(v)}\bm{\mathsf{W}}_{iv}}{\sum_{v}{e^{a^{\gamma}_{v}}\bm{\mathsf{W}}_{iv}}},
\label{eq:fusion}
\end{equation}
where $a_{v}$ is a learnable scalar weight of $v$-th view and $\gamma$ is a adjustment factor. Apparently simple as Eq. (\ref{eq:fusion}) seems, it serves two purposes: i) Distinct from other methods to treat all views equally, we flexibly assign each view different weighting coefficients, which help to maintain or highlight the differences of discriminative ability among views.
ii) Unusable views must be ignored in multi-view fusion to avoid negative effects, so missing-view indicator matrix is introduced in our fusion module. Stack all $\bar{z_i}$ and we can get output tensor $\bar{\bm{\mathsf{Z}}}\in \mathbb{R}^{n\times d_e}$ for next classification.
\subsection{CFormer and Classifier}
In the real world, multiple categories of samples are not independent of each other. How to leverage the multi-label correlations to make our model more discriminative is the main concern in the subsection. Instead of learning a category correlation graph, similar to \cite{lanchantin2021general}, we directly map each category to the feature space and leverage the self-attention mechanism to capture the category correlations. In more detail, $c$ class tokens $\bigl\{cls^i\in \mathbb{R}^{d_e}\bigr\}_{i=1}^{c}$ are randomly initialized before training and input to category-aware transformer (\textbf{CFormer} for short) with fusion features of samples (\textit{i.e.}, the input of CFormer is $\{\bar{z_i},cls^1,...,cls^c\}_{i=1}^{n}$). Similar to the structure of VFormer, CFormer allows the fusion features and category tokens to share information, which enjoys two major benefits: On the one hand, the view-fusion features aggregate all subcategory information according to similarities among them, which makes the embedding features encoded by CFormer closer to their relevant class tokens. On the other hand, the information interaction across categories implicitly promotes the learning of category relevance. It should be pointed out that we do not introduce the missing-label mask here because mining the category correlation information requires the participation of all class tokens. And for any sample $i$, the output tensor of our CFormer is $\{\widehat{z_i},cls^1_{i},...,cls^c_{i}\}\in \mathbb{R}^{(c+1)\times d_e}$. We split the output into two parts, \textit{i.e.}, consensus representation $\{\widehat{z_i}\}$ for the main classification purpose and $\{cls^1_{i},...,cls^c_{i}\}$ for the representation learning of class tokens. 

As shown in Fig. \ref{fig:main}, $c+1$ linear classifiers $\{\bm{\varPsi}_z,\{\bm{\varPsi}_i\}_{i=1}^{c}\}$ are connected in parallel to the outputs of CFormer, where the classifier $\bm{\varPsi}_z$ predicts the final result $\bm{\mathsf{P}}_z\in [0,1]^{n\times c}$. While each other classifier $\bm{\varPsi}_i$ only predicts the value of its own category so we can get another prediction $\bm{\mathsf{P}}_c\in [0,1]^{n\times c}$ from all class tokens. Finally, we define a masked binary cross-entropy function as our multi-label classification loss:
\begin{equation}
\small
\begin{aligned}
\bm{\mathcal{L}}_{mbce} &=  -\frac{1}{C}\sum\limits_{i=1}^{n}\sum\limits_{j=1}^{c}\Big(\bm{\mathsf{Y}}_{ij}\log(\bm{\mathsf{P}}_{ij})\\
&+(1-\bm{\mathsf{Y}}_{ij})\log(1-\bm{\mathsf{P}}_{ij})
\Big)\bm{\mathsf{G}}_{ij},
\end{aligned}
\end{equation}
where $C = \sum_{i,j}\bm{\mathsf{G}_{i,j}}$ denotes the number of available labels and $\bm{\mathsf{P}}$ is the prediction. $\bm{\mathsf{G}}$ is introduced to prevent unknown labels from participating in the calculation of loss. As a result, we can obtain a main classification loss $\bm{\mathcal{L}}_{mc}$ and an ancillary classification loss $\bm{\mathcal{L}}_{ac}$ according to $\bm{\mathcal{L}}_{mbce}$ for $\bm{\mathsf{P}}_z$ and $\bm{\mathsf{P}}_c$, respectively. Overall, our total loss function is :
\begin{equation}
\small
\label{eq:loss}
\bm{\mathcal{L}} = \bm{\mathcal{L}}_{mc}+ \alpha\bm{\mathcal{L}}_{gc}+ \beta\bm{\mathcal{L}}_{ac},
\end{equation} 
where $\alpha$ and $\beta$ are penalty coefficients.

\section{Experiments}
\subsection{Experimental Setting}
\textbf{Datasets.} In the experiments, we use five multi-view multi-label datasets as same as \cite{tan2018incomplete,guillaumin2010multimodal,li2020concise}: (1) \textit{corel5k} \cite{duygulu2002object}: The corel5k dataset contains 5000 image samples with 260 classes and we use 4999 samples in the experiments. (2) \textit{Pascal07} \cite{everingham2009pascal}: The popular Pascal07 dataset has 9963 images covering 20 types of tags. (3) \textit{Espgame} \cite{von2004labeling}: 20770 samples and 268 categories of Espgame are used in our experiments. It is no doubt that it is a large-scale database. (4) \textit{IAPRTC12} \cite{grubinger2006iapr}: As a benchmark database, IAPRTC12 is composed of 20,000 high-quality natural images and we use 19627 images and 291 tags in the experiments. (5) \textit{Mirflickr} \cite{huiskes2008mir}: The Mirflickr is collected from the social photography site Flickr, which is made up by 25000 images. 38 types of annotations are selected in the experiments. For the above five multi-view multi-label datasets, each dataset contains six views, \textit{i.e.}, GIST, HSV, HUE, SIFT, RGB, and LAB. 

\textbf{Data processing.} To evaluate the performance of various multi-view multi-label learning methods on incomplete datasets, following \cite{tan2018incomplete}, we treat the five datasets as follows to simulate the incomplete case: (1) For each view, we randomly remove $50\%$ of samples while guaranteeing that at least one view is available for each sample. (2) For each category, we randomly select $50\%$ of positive tags and negative tags as unknown labels. 

\textbf{Comparison.} We select eight top methods to compare to our LMVCAT. Six methods, \textit{i.e.}, lrMMC, MVL-IV, MvEL, iMSF, iMvWL, and NAIML, are introduced in section \ref{2.2}. In addition, we add two advanced methods, named C2AE \cite{yeh2017learning} and GLOCAL \cite{zhu2017multi}, to expand the evaluation experiments. C2AE is a canonical correlated autoencoder network, which learns a latent embedding space to bridge feature representations and label information. GLOCAL exploits the global and local label correlations on the representation learning, which is also an application of label manifold. However, C2AE and GLOCAL are both single-view multi-label classification methods, and only iMvWL and NAIML are designed for the datasets with both missing views and missing labels. Therefore, we have to do extra alterations to the rest of the methods. Like \cite{tan2018incomplete} and \cite{li2020concise}, average values of available views are filled into unusable views for MvEL and lrMMC. And as to MVL-IV and iMSF, we set missing tags to be negative tags. C2AE and GLOCAL are independently conducted on each view and the best results are reported. All parameters of these comparison methods are set as recommended in their papers or codes for a fair comparison.

\textbf{Evaluation.} Different from \cite{tan2018incomplete} and \cite{li2020concise}, we only select AP (Average Precision), RL (Ranking Loss), and AUC (adapted Area Under Curve) as our metrics due to the weak discrimination of HL (Hamming Loss). Note that 1-RL is used instead of RL in our results so that the higher the value, the better the performance. 

The implementation of our method is based on MindSpore and Pytorch framework.
\begin{table*}[t!]

\begin{small}
\begin{center}
    \begin{tabular}{ccccccccccc}
   	\toprule[1pt]
    Data  & lrMMC   & MVL-IV   & MvEL   & iMSF   & C2AE & GLOCAL & iMvWL & NAIML & ours\\
    \midrule
    Corel5k
	&.240(.002)	&.240(.001)	&.204(.002)	&.189(.002)	& .227(.008) & 0.285(0.004) &.283(.007)	&.309(.004) &\textbf{.382(.004)}\\
	Pascal07
	&.425(.003)	&.433(.002)	&.358(.003)	&.325(.000)	& .485(.008) & 0.496(0.004) &.441(.017)	&.488(.003)	 &\textbf{.519(.005)}\\
	Espgame
	&.188(.000)	&.189(.000)	&.132(.000)	&.108(.000)	&.202(.006) &0.221(0.002) &.242(.003)	&.246(.002)	 &\textbf{.294(.004)}\\
	Iaprtc12
	&.197(.000)	&.198(.000)	&.141(.000)	&.101(.000)	& .224(.007) & 0.256(0.002) &.235(.004)	&.261(.001)	 &\textbf{.317(.003)}\\
	Mirflickr
	&.441(.001)	&.449(.001)	&.375(.000)	&.323(.000)	& .505(.008) & 0.537(0.002) &.495(.012)	&.551(.002) &\textbf{.594(.005)}\\
	\bottomrule[1pt]
\end{tabular}
\end{center}
\end{small}
\caption{The AP values of nine methods on the five datasets with 50\% missing-view ratio, 70\% training samples, and 50\% missing-label ratio. The best resluts are marked in bold.}
\label{table:ap}
\end{table*}

\begin{table*}[t!]

\begin{small}
\begin{center}
    \begin{tabular}{ccccccccccc}
   	\toprule[1pt]
    Data  & lrMMC   & MVL-IV   & MvEL   & iMSF   & C2AE & GLOCAL & iMvWL & NAIML & ours\\
    \midrule
    Corel5k
	&.762(.002)	&.756(.001)	&.638(.003)	&.709(.005) &.804(.010) &0.840(0.003)	&.865(.003)	&.878(.002)	&\textbf{.880(.002)}\\
	Pascal07
	&.698(.003)	&.702(.001)	&.643(.004) &.568(.000) &.745(.009) &0.767(0.004)	&.737(.009)	&.783(.001) &\textbf{.811(.004)}\\
	Espgame
	&.777(.001)	&.778(.000)	&.683(.002)	&.722(.002) &.772(.006) &0.780(0.004)	&.807(.001)	&.818(.002)	&\textbf{.828(.002)}\\
	Iaprtc12
	&.801(.000)	&.799(.001)	&.725(.001)	&.631(.000) &.806(.005) &0.825(0.002)	&.833(.003)	&.848(.001) &\textbf{.870(.001)}\\
	Mirflickr
	&.805(.000)	&.804(.001)	&.746(.001)	&.665(.001) &.821(.003) &0.832(0.001)	&.836(.002)	&.850(.001)	&\textbf{.865(.003)}\\
	\bottomrule[1pt]
\end{tabular}
\end{center}
\end{small}
\caption{The 1-RL values of nine methods on the five datasets with 50\% missing-view ratio, 70\% training samples, and 50\% missing-label ratio. The best resluts are marked in bold.}
\label{table:rl}
\end{table*}

\begin{table*}[t!]

\small
\begin{center}
    \begin{tabular}{ccccccccccc}
   	\toprule[1pt]
    Data  & lrMMC   & MVL-IV   & MvEL   & iMSF   & C2AE & GLOCAL & iMvWL & NAIML & ours\\
    \midrule
    Corel5k
	&.763(.002)	&.762(.001)	&.715(.001)	&.663(.005) & .806(.010) & 0.843(0.003)	&.868(.003)	&.881(.002)	&\textbf{.883(.002)}\\
	Pascal07
	&.728(.002)	&.730(.001)	&.686(.005)	&.620(.001) & .765(.010) & 0.786(0.003)	&.767(.012)	&.811(.001)	&\textbf{.834(.004)}\\
	Espgame
	&.783(.001)	&.784(.000)	&.734(.001)	&.674(.003) & .777(.006) & 0.784(0.004)	&.813(.002)	&.824(.002)	&\textbf{.833(.002)}\\
	Iaprtc12
	&.805(.000)	&.804(.001)	&.746(.001)	&.665(.001) & .807(.005) & 0.830(0.001)	&.836(.002)	&.850(.001)	&\textbf{.872(.001)}\\
	Mirflickr
	&.806(.001)	&.807(.000)	&.761(.000)	&.715(.001) & .810(.004) & 0.828(0.001)	&.794(.015)	&.837(.001)	&\textbf{.853(.003)}\\
	\bottomrule[1pt]
\end{tabular}
\end{center}

\caption{The AUC values of nine methods on the five datasets with 50\% missing-view ratio, 70\% training samples, and 50\% missing-label ratio. The best resluts are marked in bold.}
\label{table:auc}
\end{table*}

\subsection{Experimental Results and Analysis}

\begin{figure}[h!]
		\centering
		\subfloat[different missing-view ratios]{

			\includegraphics[width=0.48\linewidth,height=2.7cm]{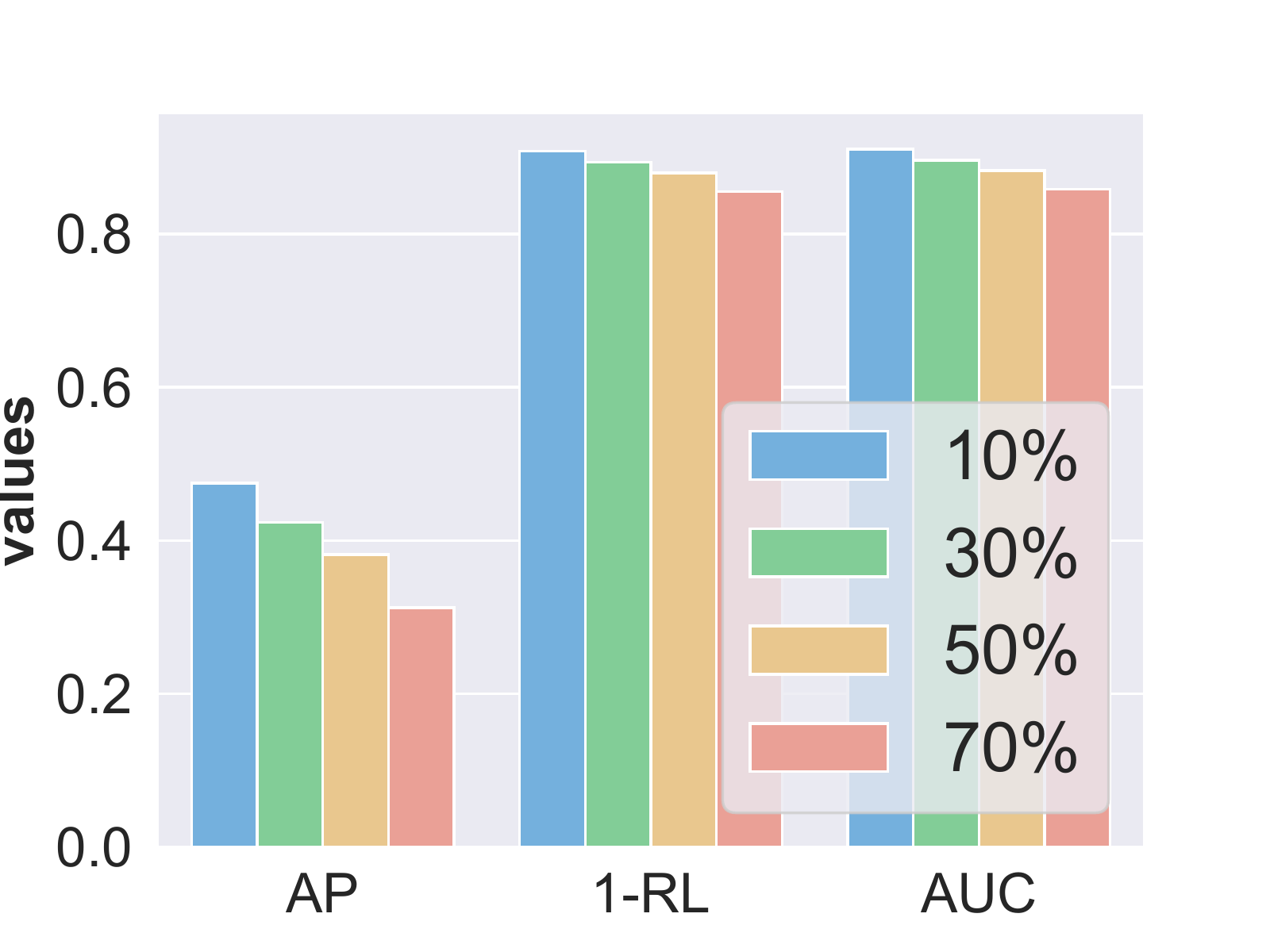}
		}
		\subfloat[different missing-label ratios]{

			\includegraphics[width=0.48\linewidth,height=2.7cm]{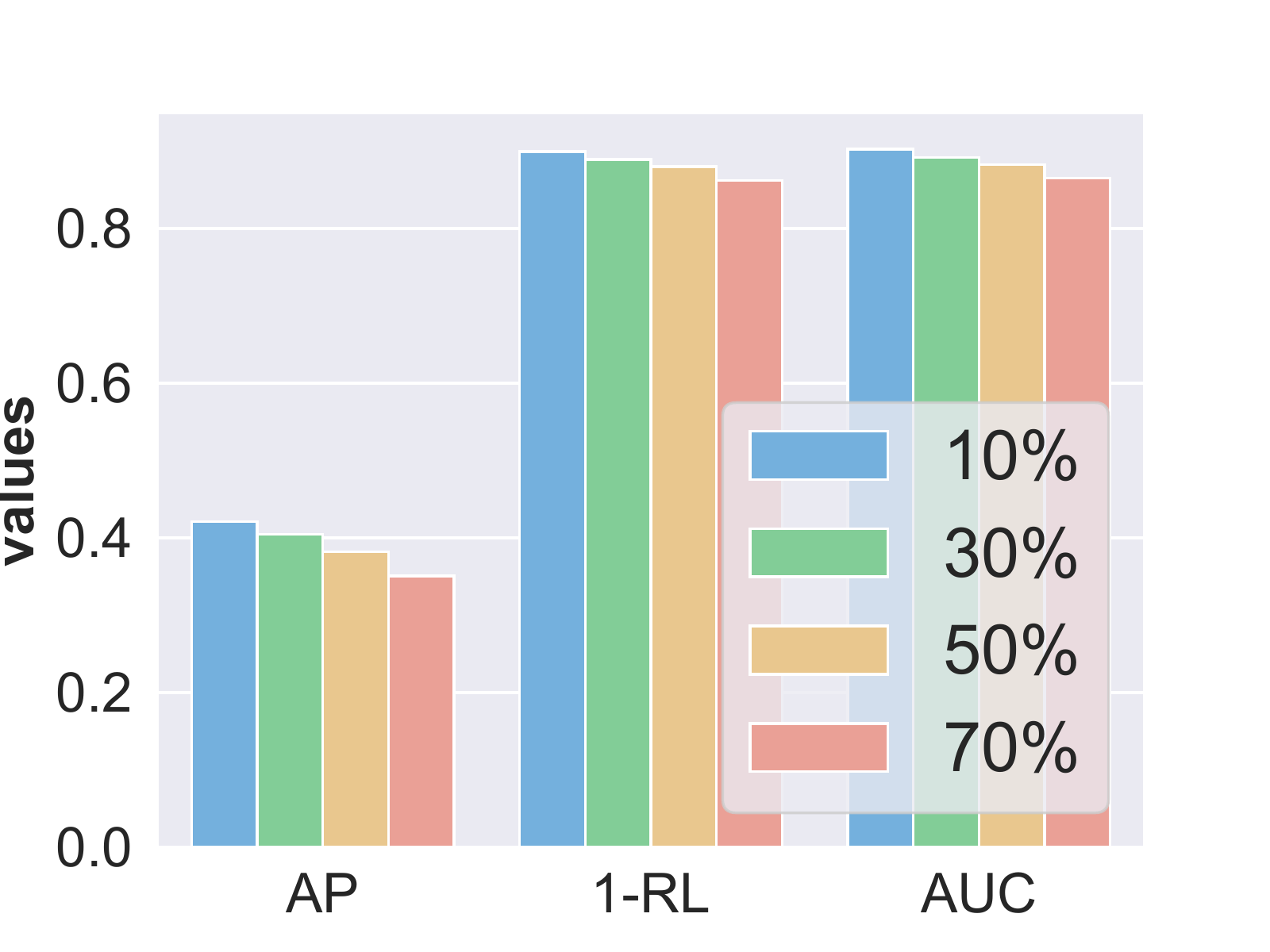}
		}
		
		\caption{The results on Corel5k dataset with (a) different missing-view ratios and a 50\% missing-label ratio and (b) a 50\% missing-view ratio and different missing-label ratios.}
		\label{fig:miss-rates}
\end{figure}
Table \ref{table:ap} to \ref{table:auc} show the performance of the nine methods on five incomplete datasets (some of the results are quoted from \cite{li2020concise} and \cite{tan2018incomplete}). Table \ref{table:exp2} lists the results on the datasets with complete views and labels. All experiments are repeated 10 times to avoid outlier results as much as possible, and we list the mean and standard deviation of 10 tests in these tables. Values in parentheses represent the standard deviation. Fig. \ref{fig:miss-rates} and Fig. \ref{fig:samples-rates} show more results \textit{w.r.t} different missing and training sample rates. From Table \ref{table:ap} to \ref{table:exp2}, we can summarize the following observations:
\begin{figure}[t!]
		\centering
		\subfloat[Corel5k]{

			\includegraphics[width=0.48\linewidth,height=2.7cm]{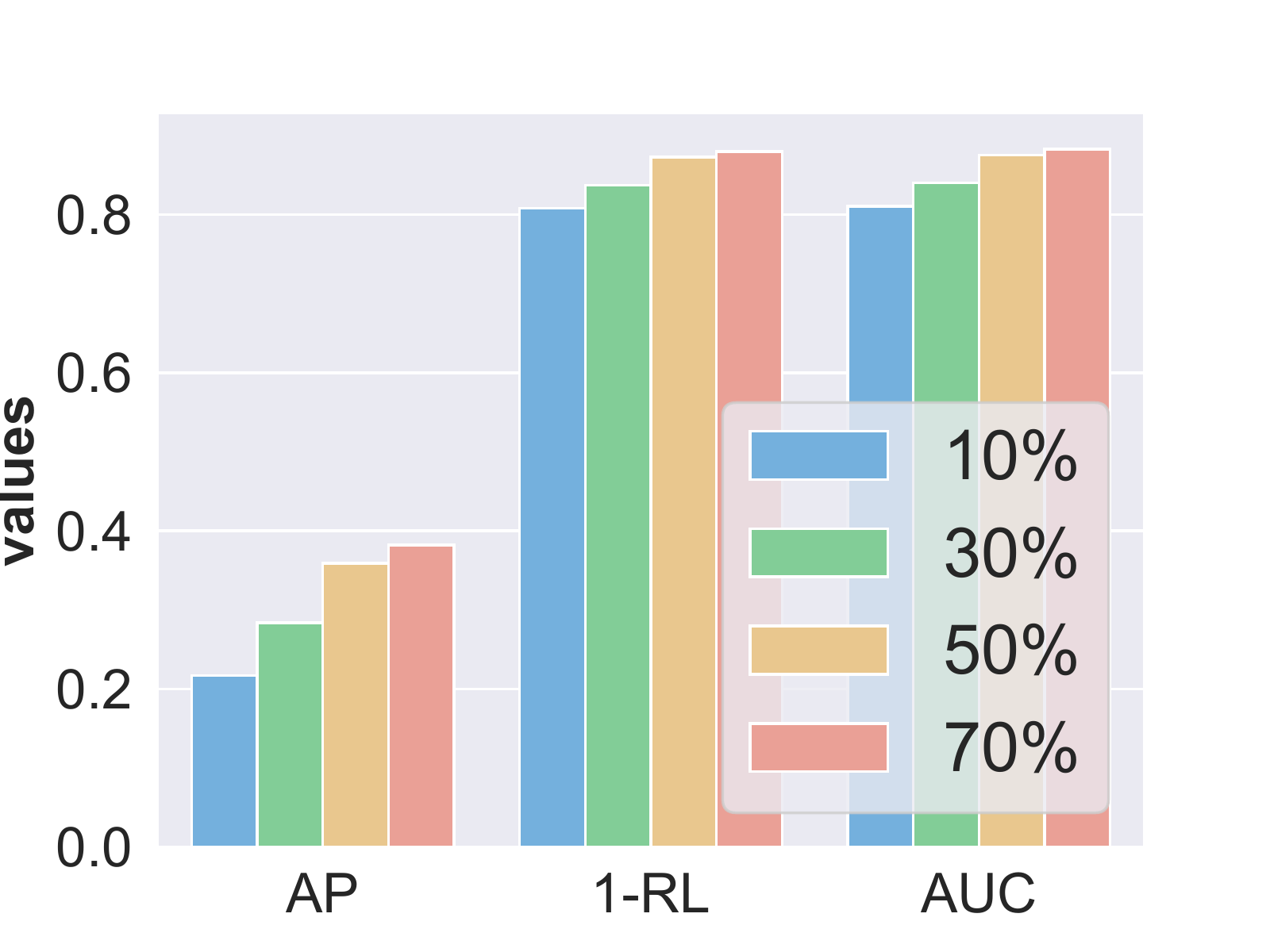}
		}
		\subfloat[Mirflickr]{

			\includegraphics[width=0.48\linewidth,height=2.7cm]{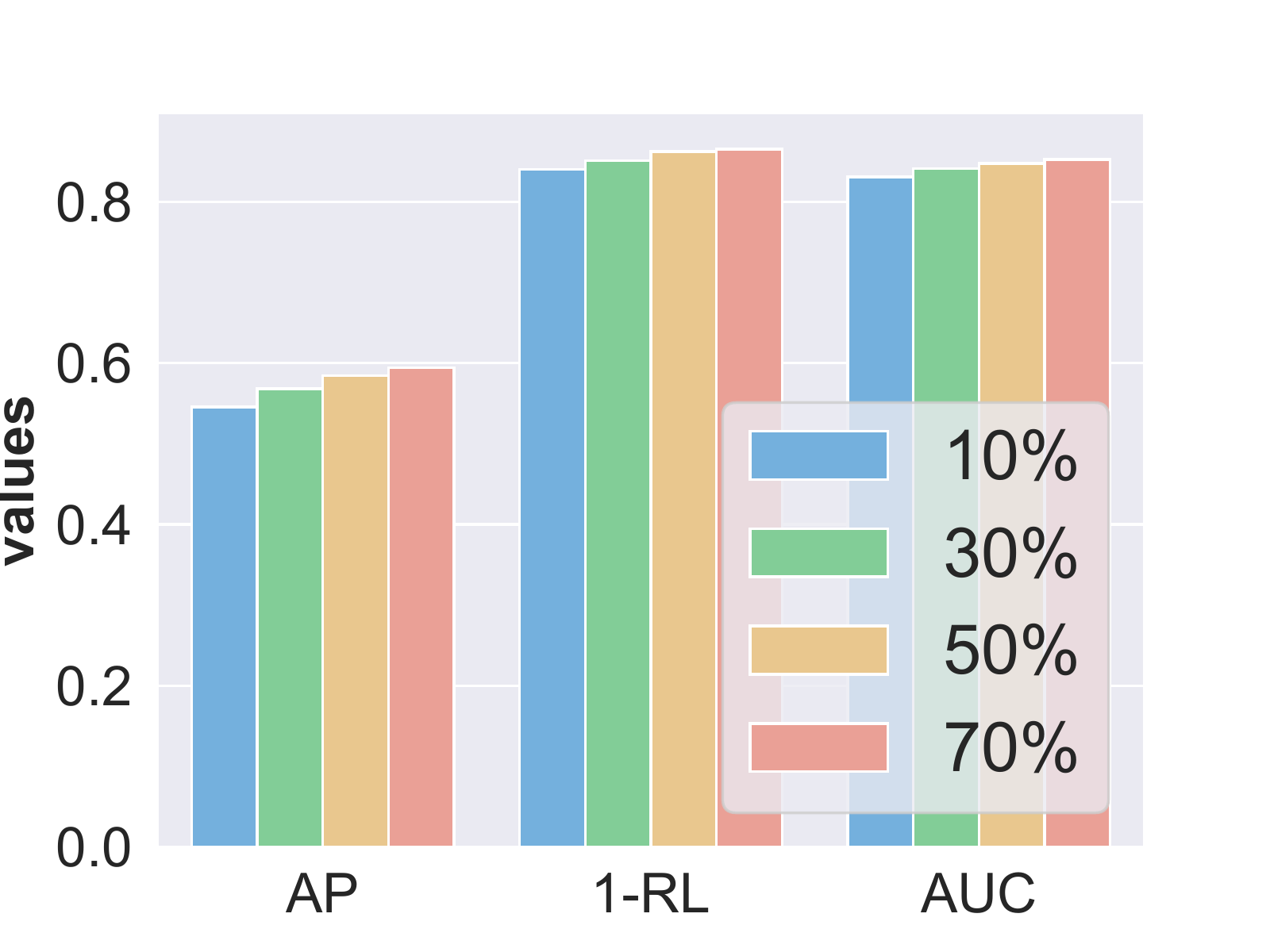}
		}
		\caption{The results on (a) Corel5k dataset and (b) Mirflickr dataset with a 50\% missing-view ratio, a 50\% missing-label ratio, and different training sample ratios.}
		\label{fig:samples-rates}
\end{figure}
\begin{itemize}
\item Our method achieves overwhelming advantages on all three metrics of the five datasets, especially on the most representative AP metric. For example, the AP value of our model exceeds the second-best NAIML by about 7\% and 5\% on the Corel5k and Espgame datasets respectively, which verifies the superiority of our method. 
\item Both iMvWL and NAIML reach relatively better performance compared to the first four methods, which benefits from good compatibility to double incomplete data. As a deep method with stronger fitting ability, C2AE performs mediocrely due to the lack of multi-view complementary information. Though GLOCAL is a single-view method, the full exploitation of label correlations helps it achieve surprising results on several datasets.
\item From Table \ref{table:exp2}, we can find that, although our LMVCAT is an incomplete method, it still works well with complete datasets. Specifically, on the Corel5k dataset, the AP value of our method is about 14 percentage points ahead of the second-best GLOCAL.
\end{itemize}
\begin{table}[t!]
\begin{small}
\tabcolsep=1.2mm
\begin{center}
    \begin{tabular}{c|c|ccccc}
   	\toprule[1pt]

    Dateset & Metric & C2AE & GLOCAL & iMvWL & NAIML & ours\\
    \midrule
    \multirow{3}[1]{*}{\begin{turn}{0}Corel5k\end{turn}}
		&AP	&.353	&.386 	&.313	&.327 &\textbf{.521}\\
		&1-RL	&.870	&.891	&.884	&.890 &\textbf{.928}\\
		&AUC	&.873	&.895	&.887	&.893 &\textbf{.930}\\
	\midrule
	 \multirow{3}[1]{*}{\begin{turn}{0}Pascal07\end{turn}}
		&AP	&.584	&.610	&.468	&.496 &\textbf{.629}\\
		&1-RL	&.835	&.866 	&.763	&.795 &\textbf{.878}\\
		&AUC	&.851	&.879 	&.793	&.822 &\textbf{.892}\\
	\midrule
	 \multirow{3}[1]{*}{\begin{turn}{0}Espgame\end{turn}}
		&AP	&.269	&.264 	&.260	&.251 &\textbf{.385}\\
		&1-RL	&.832	&.804 	&.817	&.825 &\textbf{.876}\\
		&AUC	&.837	&.810 	&.822	&.830 &\textbf{.880}\\
	\midrule
	 \multirow{3}[1]{*}{\begin{turn}{0}Iaprtc12\end{turn}}
		&AP	&.316	&.330 	&.250	&.267 &\textbf{.436}\\
		&1-RL	&.868	&.871	&.842	&.825 &\textbf{.918}\\
		&AUC	&.869	&.877 	&.843	&.830 &\textbf{.918}\\
	\midrule
	\multirow{3}[1]{*}{\begin{turn}{0}Mirflickr\end{turn}}
		&AP	&.593	&.643 	&.493	&.555 &\textbf{.684}\\
		&1-RL	&.854	&.876 	&.806	&.847 &\textbf{.905}\\
		&AUC	&.845	&.869 	&.789	&.839 &\textbf{.889}\\
	\bottomrule[1pt]
\end{tabular}
\end{center}
\end{small}
\caption{The experimental results on the five datasets with full views, full labels and 70\% training samples. The best results are marked in bold.}
\label{table:exp2}
\end{table}
\begin{figure}[t!]
		\centering
		\subfloat[Corel5k]{

			\includegraphics[width=0.46\linewidth,height=3cm]{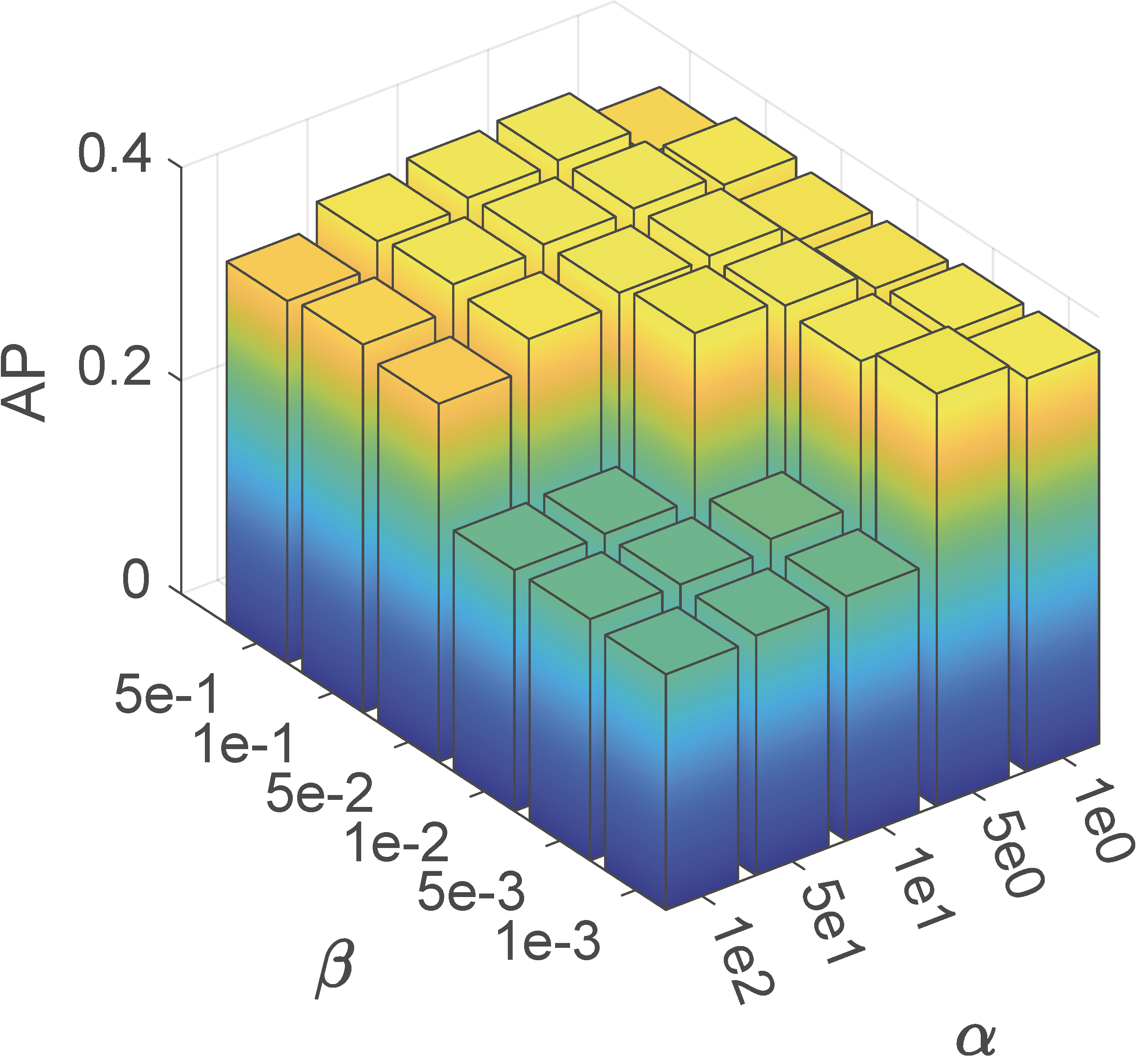}
		}
		\subfloat[Pascal07]{

			\includegraphics[width=0.46\linewidth,height=3cm]{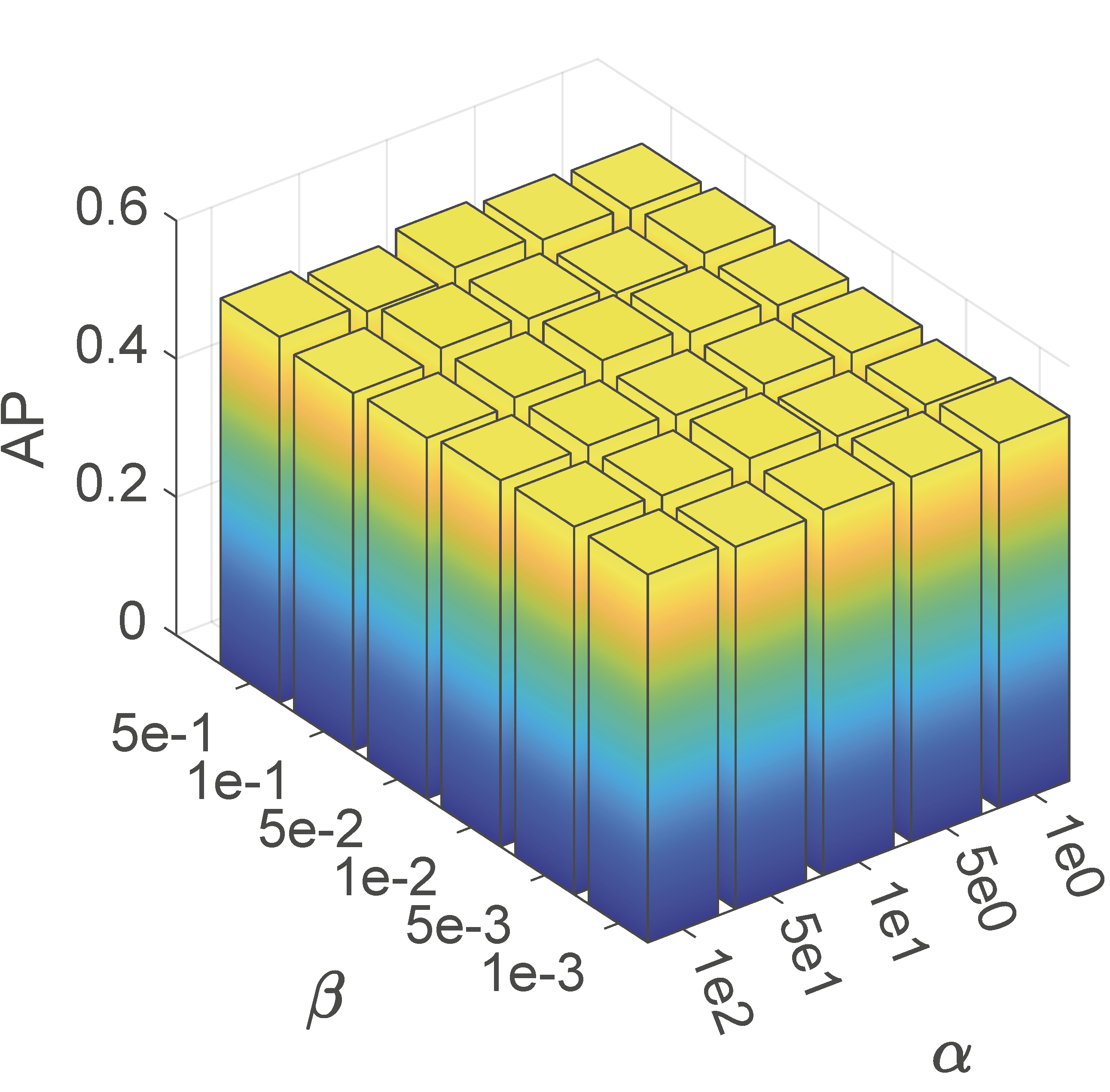}
		}
		\quad
		\subfloat[Corel5k]{
		
			\includegraphics[width=0.48\linewidth,height=2.7cm]{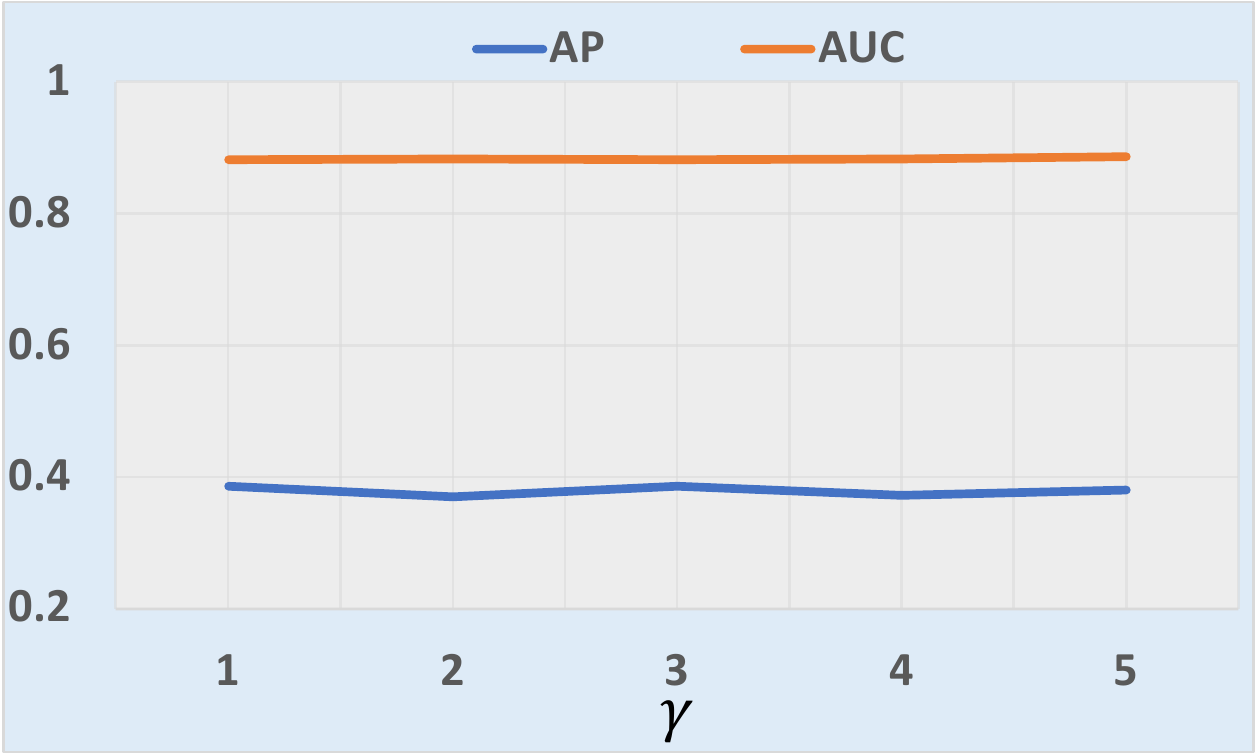}
		}
		\subfloat[Pascal07]{

			\includegraphics[width=0.48\linewidth,height=2.7cm]{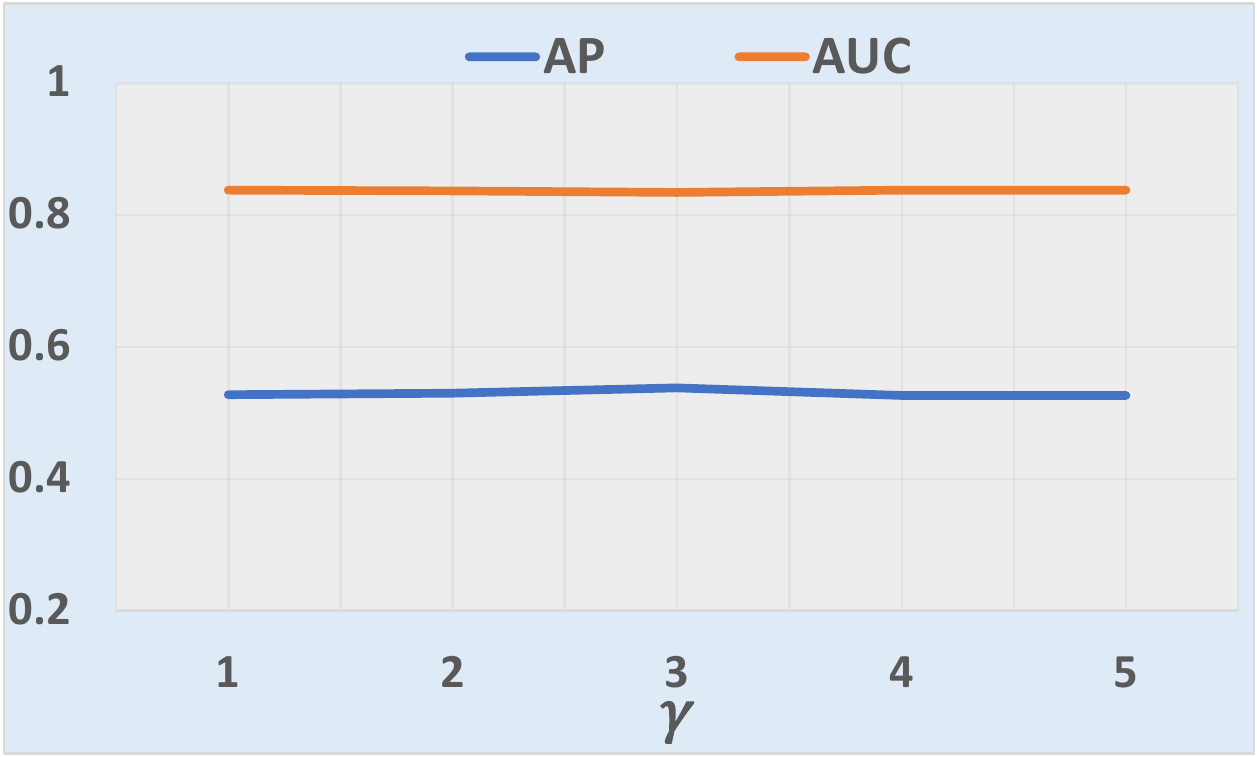}
		}
		\caption{The results of different parameter selections on (a,c) Corel5k and (b,d) Pascal07 datasets with half the available views and known labels and a 70\% training sample rate.}
		\label{fig:param}
\end{figure}

As shown in Fig. \ref{fig:miss-rates}, we respectively plot the histogram of three metrics in different missing-view and missing-label ratios by fixing another ratio on the corel5k dataset. These results verify that both incomplete views and partial labels are harmful for efficient classification. In addition, an interesting phenomenon is that when the incompleteness ratio is relatively small, missing labels have a greater impact on the performance, and conversely, the negative impact of missing views is greater. 
As shown in Fig. \ref{fig:samples-rates}, we conduct experiments on Corel5k and Mirflickr datasets with 50\% available views, 50\% known tags and different training sample ratios. It is clear that, as the proportion of training samples in the total increases, the performance of our method gradually goes up. 

\subsection{Hyperparameters Study}
There are three hyperparameters, \textit{i.e.}, $\alpha$, $\beta$, and $\gamma$ in our LMVCAT. To study the optimal parameters selection, we list the performance of our method on different parameter combinations in Fig. \ref{fig:param}. All datasets used in parameters study are with 50\% missing-view ratio, 50\% missing-label ratio, and 70\% training samples. As can be seen in Fig. \ref{fig:param} (a) and (b), for Corel5k dataset, the optimal parameters $\alpha$ and $\beta$ are located in the range of $[5,10]$ and $[0.05,0.5]$, respectively, and for Pascal07 dataset, the optimal parameters $\alpha$ and $\beta$ are located in the range of $[10,100]$ and $[0.001,0.5]$, respectively. As to $\gamma$, obviously, our method is insensitive to it from Fig. \ref{fig:param} (c) and (d). In our experiments, $\gamma$ is uniformly set to 2.
\begin{table}[t!]
\begin{small}
\tabcolsep=1.2mm
\begin{center}
    \begin{tabular}{c|ccc|ccc}
   	\toprule[1pt]
	    \multirow{2}{*}{method}  & \multicolumn{3}{c|}{Corel5k} & \multicolumn{3}{c}{Espgame}\\
     & AP   & 1-RL   & AUC & AP   & 1-RL   & AUC  \\
    \midrule
    $\mathcal{V}$
	&.347	&.871	&.874	&.284	&.822	&.827\\	
	$\mathcal{V}+\mathcal{A}$
	&.348	&.872	&.877	&.286	&.824	&.829\\
	$\mathcal{V}+\mathcal{A}+\mathcal{C}$
	&.359	&.879	&.883	&.288	&.828	&.833\\
	$\bm{\mathcal{V}+\mathcal{A}+\mathcal{C}+\mathcal{G}}$
	&\textbf{.382}	&\textbf{.880}	&\textbf{.883}	&\textbf{.294}	&\textbf{.828} &\textbf{.833}\\
	\textit{w/o} view mask
	&.352	&.866	&.870	&.274	&.817 &.823\\
	\textit{w/o} label mask
	&.365	&.880	&.883	&.287	&.827 &.832\\
	\bottomrule[1pt]
\end{tabular}
\end{center}
\end{small}
\caption{The ablation experiments on two datasets with 50\% missing-view ratio, 50\% missing-label ratio, and 70\% training samples. $\mathcal{V}$ is the backbone with VFormer; $\mathcal{A}$ denotes the adaptively weighted strategy; $\mathcal{C}$ represents the CFormer; and $\mathcal{G}$ means the graph constraint.}
\label{table:ablation}
\end{table}

\subsection{Ablation Study}
To study the effectiveness of each component of our method, we perform experiments with following altered methods. For the complete LMVCAT, we remove graph constraint, CFormer, and adaptively weighted module sequentially. Besides, we take off the missing-view indicator and missing-label indicator in our model so that the incomplete views and weak labels are completely exposed to the network. As can be seen in Table \ref{table:ablation}, our label-guided graph constraint plays a key role and all experiments confirm that each component of our LMVCAT is beneficial and necessary.

\section{Conclusion} In this paper, we propose a general transformer-based framework for \textit{MvMlC}, which skillfully exploits label manifold to guide the representation learning. And its innovative view- and category-awareness realize multi-view information interaction and multi-category discrimination fusion. An adaptively weighted fusion strategy is also introduced to balance the view-specific contribution. In addition, missing-view and weak label problems are specifically avoided throughout the network. Extensive experiments support the conclusion of the effectiveness of our method.

\section*{Acknowledgments}
This work is supported by Shenzhen Science and Technology Program under Grant RCBS20210609103709020, GJHZ20210705141812038, Shenzhen Fundamental Research Fund under Grant GXWD20220811173317002, and CAAI-Huawei MindSpore Open Fund under Grant CAAIXSJLJJ-2022-011C.

\bibliography{aaai23}



\end{document}